%% file: main.tex
\documentclass[twocolumn,a4wide]{article}

\usepackage{geometry}
\usepackage[utf8]{inputenc} %
\usepackage[T1]{fontenc}    %
\usepackage[colorlinks]{hyperref}
\usepackage{xurl}           %
\usepackage{booktabs}       %
\usepackage{amsfonts}       %
\usepackage{graphicx}
\usepackage{color}
\usepackage{amssymb}
\usepackage{amsmath}
\usepackage{xspace}
\usepackage{nicefrac}       %
\usepackage{microtype}      %
\usepackage{xcolor}         %
\usepackage{ulem}
\usepackage{multirow}
\usepackage{soul}
\usepackage{colortbl}
\usepackage{multicol}

\usepackage{palatino}

\usepackage{textcomp}%
\usepackage{mathtools,amssymb,amsthm} %
\usepackage[libertine,cmintegrals,cmbraces,vvarbb]{newtxmath}
\usepackage[scr=boondoxo]{mathalfa}%
\usepackage{bm}%

\renewcommand{\emph}[1]{\textit{#1}}

\title{Vector search with small radiuses}

\author{
Gergely Szilvasy
\and
Pierre-Emmanuel Mazar\'e
\and
Matthijs Douze
}
\date{}

\def\eg{\textit{e.g.~}}
\def\ie{\textit{i.e.~}}

\begin{document}
\maketitle

\begin{abstract}
In recent years, the dominant accuracy metric for vector search is the recall of a result list of fixed size (top-k retrieval), considering as ground truth the exact vector retrieval results. 
Although convenient to compute, this metric is distantly related to the end-to-end accuracy of a full system that integrates vector search. 
In this paper we focus on the common case where a hard decision needs to be taken depending on the vector retrieval results, for example, deciding whether a query image matches a database image or not. 
We solve this as a range search task, where all vectors within a certain radius from the query are returned. 

We show that the value of a range search result can be modeled rigorously based on the query-to-vector distance. 
This yields a metric for range search, RSM, that is both principled and easy to compute without running an end-to-end evaluation.
We apply this metric to the case of image retrieval. 
We show that indexing methods that are adapted for top-k retrieval do not necessarily maximize the RSM. 
In particular, for inverted file based indexes, we show that visiting a limited set of clusters and encoding vectors compactly yields near optimal results. 

\end{abstract}

\section{Introduction}

\input{1-intro.tex}

\section{Related work}
\label{sec:related}
\input{2-related.tex}

\section{The RSM evaluation metric}
\label{sec:metric}
\input{3-metric.tex}

\section{Evaluation on vector indexes}
\label{sec:method}
\input{4-method.tex}

\section{Conclusion}
\label{sec:conclusion}

In this work we derived the RSM metric to evaluate range search on vector databases for applications that perform a post-verification. 
Using this metric, we revisit some vector search  principles and find that it is less useful to explore deep for non-exhaustive search, and that short encodings work almost as well as long as long ones.

One interesting observation is that the training of SSCD descriptors that this study is based on relies heavily on the “catalyzer loss” from~\cite{sablayrolles2019spreading} for similarity search. One of the properties of this loss is to make the results of k-NN search and range more similar~\cite[appendix C]{sablayrolles2019spreading}. 
This loss is useful to include at training time for embeddings that should be retrieved with range search to improve the distance normalization.

\bibliographystyle{plain}
\bibliography{biblio}

\end{document}

%% file: 1-intro.tex
Similarity search is a task of information retrieval widely used in various applications, including ranking, recommendations, or trust and safety. 
Given a query item, the objective is to find the most similar item from a collection (the database). 
The items can be images, videos, text, products, social media posts, etc.
To enable efficient search, both the database entities and the queries can be encoded as high-dimensional vectors, typically neural embeddings~\cite{pizzi2022sscd,caron2020unsupervised,devlin2018bert}. 
Instead of exhaustively comparing queries with the entire database, we employ approximate nearest neighbor search algorithms, that strike a balance between computational cost (time and space) and search accuracy. 

In the classical web search setting, a user submits a query to the system, which returns a ranked list of results of fixed size. 
The results are then displayed to the user by decreasing order of relevance with a fixed number of results. 
In the embedding space, this is implemented by retrieving a fixed number of results per query vector: it is $k$-nearest-neighbor search (k-NN). 

However, for real-world collections of items, most query items will have no match while others may have millions. 
Besides, embedding extractors are commonly designed so that a distance threshold can distinguish matching from non-matching item pairs. 
Therefore, the relevant search criterion, rather than k-NN search, is to select all database items that are closer than a threshold to the query vector: this is range search. 
The range search results in embedding space undergo a post-verification stage: the pairs of query-database item are analyzed with a more computationally demanding method to distinguish correct from incorrect matches. 
For example, for image retrieval, the post-verification may consist in analyzing the images to do pairwise matching. 
There is a limited post-processing budget. 

Therefore, in this work, we consider a setting different from k-NN. 
The retrieval application does bulk processing, \ie a large number of queries are processed at once. 
The range search threshold is set to yield a number of pairs corresponding to the post-verification budget. 

Evaluating range search is delicate.
For most embedding distributions, the number of ground truth neighbors varies a lot from one query vector to another, to the point that most queries have no neighbor at all. 
This is partly because the number of matches for the items are imbalanced themselves, but also because the embedding vector distributions tend to be of varying density depending on the neighborhood~\cite{jegou2007contextual}.
Therefore, range search relies on a subset of the query points. 
Moreover, the distances to most ground truth neighbors are near the range search boundary, which makes the set of neighbors sensitive to low levels of noise. 

We derive a principled metric, RSM, to benchmark range search for bulk search. 
RSM is based on a very simple model of the filtering ability of the post-processing method. 
RSM is fast to evaluate and unaffected by the boundary effects of range search.

Practical vector search methods rely on non-exhaustive search to drive down the search time at some cost in accuracy. 
For example, the IVF method partitions the dataset at building time and visits only a subset of partitions at search time. 
Besides, to reduce the memory consumption of datasets, the embedding vectors are often compressed to binary vectors or other forms of quantization.
These techniques are well studied for k-NN search. 
Using our new RSM tool, we evaluate IVF indexes and vector compression for bulk range search. 

We find that the optimal indexes for these settings are not the same as for the classical k-NN. 
Useful range search results tend to be near the query vectors. 
Therefore, for non-exhaustive indexes, exploring less promising clusters of database vectors quickly yields diminishing returns. 
Similarly, using accurate vector representations that consume more memory have less impact than for k-NN search. 
In fact, binary representations compared with Hamming distances, that are fast but much less accurate than quantization based approaches, are relatively competitive in a range search setting. 

The paper starts by reviewing related work in Section~\ref{sec:related} and introduces the dataset used throughout for experiments (Section~\ref{src:dataset}). 
In section~\ref{sec:metric}, we justify and derive the RSM metric. 
Section~\ref{sec:method} applies this metric to vector search experiments, and we conclude with a summary of our key observations in Section~\ref{sec:conclusion}.

%% file: 2-related.tex
\paragraph{Vector search techniques} 

Vector search is a simple operation that can be computed exactly. 
However, at scale, most vector search libraries rely on a combination of non-exhaustive search and compression to compute an approximate result and spare computing resources~\cite{douze2024faiss}.
In this work we focus primarily on inverted-file based methods~\cite{jegou11pq,baranchuk2018revisiting} that partition the dataset and visit only a fraction of the partitions at search time. 
For compression, we evaluate binary representations~\cite{gong2013iterative,jegou2008hamming} and product quantization~\cite{jegou11pq}.

\paragraph{Vector search metrics.}

There are two broad ways of evaluating the accuracy of a vector search system, depending on the provenance of the ground-truth. 
The first evaluation is based on an end-to-end application-based metric: the vector search results are checked with a retrieval metric, typically using precision and recall~\cite{weiss2008spectral,jegou2011aggregating}.
The second evaluation compares the vector search results with an exact vector search ground truth. 
Thus, the vector search is evaluated in isolation~\cite{aumuller2020ann,simhadri22results}.
The end-to-end metric is closer to the application result that matters but the pure search metric makes it easier to perform a fine analysis of search algorithm itself, decoupled from how it is exploited~\cite{douze2024faiss}.
This work combines both advantages, since it builds a model of the end performance, without requiring to perform costly end-to-end experiments. 

Given a query vector, search APIs typically offer to find either the $k$ nearest neighbors (k-NN search) or the vectors within a certain distance $\varepsilon$ to the query (range search). 
Evaluating $k$-nearest neighbor search with a pure search metric is relatively unambiguous: both the exact search ground-truth and the system's search results are sets, so retrieval metrics like precision and recall are straightforward to compute~\cite{simhadri22results,pauleve2010locality}.
This also closely matches some applications, where the contrast of the distance to the nearest neighbor wrt. the next neighbors is a good discriminator.
This is the case, \eg for keypoint matching~\cite{Lowe04sift} or pairwise matching between two sets~\cite{conneau2017word} with contrastive neighbor distances. 

\paragraph{Evaluating range search.}
Range search is difficult to evaluate, \eg in the BigANN 2021 challenge, some ad-hoc weighting was applied to avoid that some queries would have a disproportionate impact on the metric~\cite{simhadri22results}. 
However, range search has important applications, for example to do bulk matching of image pairs~\cite{douze2021isc}. 
Besides, early vector search methods like Locality-Sensitive Hashing (LSH) and its variants are theoretically grounded on range search: the probability of hash collisions is conditional on vectors being within a certain range~\cite{indyk1998approximate,datar2004locality}.
In this work, we focus on range search and show how a principled weighting metric can be used to evaluate the relevance of matching results. 

\paragraph{Distance calibration.}

Searching a set of vectors with a fixed distance $\varepsilon$ means that the distances between true positive matching pairs should be roughly the same in the whole space. 
This property is not obvious, for some vector distributions, the distance to neighbors can be very different depending on the location of the query in the vector space~\cite{jegou2007contextual}. 
For embeddings computed by neural nets, distances can be calibrated directly in the loss.
In the popular ArcFace triplet loss~\cite{deng2019arcface}, there is a margin intended to separate positive from negative training pairs. 
The entropy loss used in the \emph{spreading vectors} work~\cite{sablayrolles2019spreading} is based on the negative logarithm of the distance between nearby points. 
Therefore, the loss becomes abruptly very high below some distance threshold. 
Even in the SimCLR unsupervised training loss~\cite{chen2020simple}, that is based on a cross-entropy loss, the distances are calibrated using the temperature parameter of the softmax. 
In this work, we assume the neighborhood distance is calibrated.

\paragraph{Multi-stage retrieval.}

Multi-stage retrieval systems combine a fast first-level retrieval step that produces a short-list of results with a slower, more accurate retrieval step that reorders or filters the short-list. 
This is typical for image search systems that are performed in two stages, with a scalable vector search first and a stricter filtering applied on a shortlist of results returned by the first-level retrieval~\cite{philbin2007object,jegou2008hamming,sun20213rd}. 
Recommender systems can also be built in multiple stages~\cite{liu2022neural}, where the first level is a vector search engine. 
In this work, we develop a model of the filtering rate of the second stage to assess the quality of the vector search.

\section{The dataset}
\label{src:dataset}

The experiments in this work operate on an image dataset whose features have been extracted with a specific CNN model. 

\newcommand{\nq}{n_\mathrm{q}}
\newcommand{\nb}{N}

\newcommand{\nmatch}{n_\mathrm{match}}
\newcommand{\ninlier}{n_\mathrm{inlier}}
\newcommand{\OURD}{\textsc{RunOfTheMill}\xspace}

\subsection{The image collection}

\newcommand{\ig}[1]{\includegraphics[width=0.42\columnwidth]{figs/img/#1.jpg}}
\newcommand{\imcreds}[1]{\raisebox{\depth}{\scalebox{0.4}{\rotatebox{270}{#1}}}}
\begin{figure}
    \centering
\begin{tabular}{cc}
\multicolumn{2}{c}{pairs with 9 matching points} \\
\ig{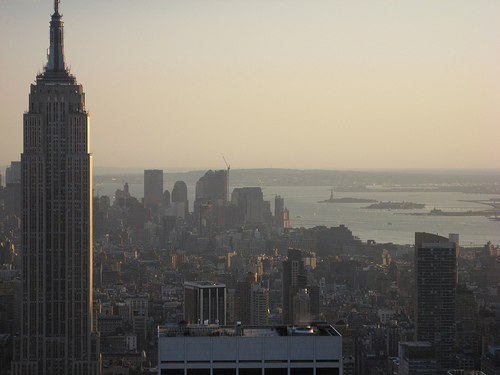}\imcreds{markagibson} & \ig{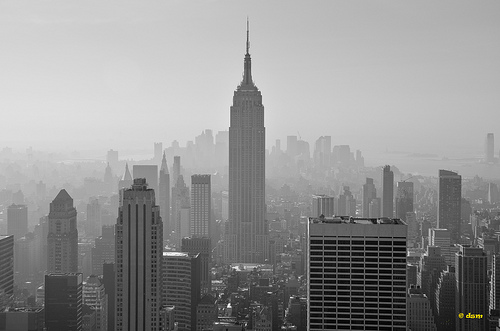}\imcreds{Davidsm72} \\
\ig{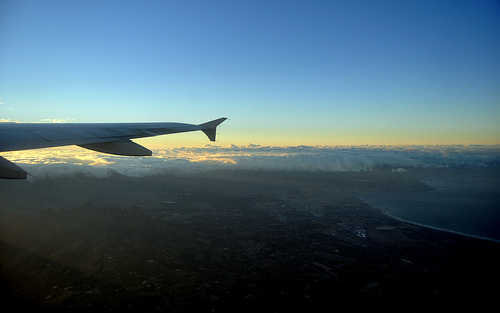}\imcreds{Meraj Chhaya} & \ig{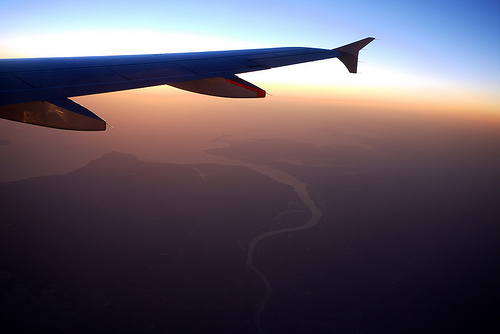}\imcreds{islandjoe} \\
\multicolumn{2}{c}{pairs with 25\% of points that match} \\
\ig{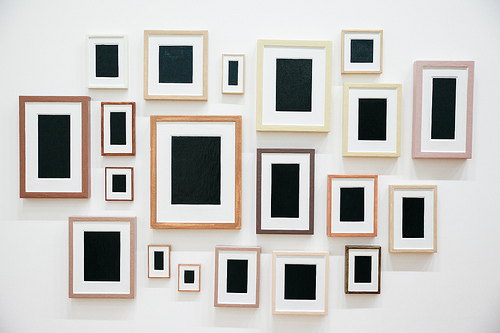}\imcreds{bortescristian} & \ig{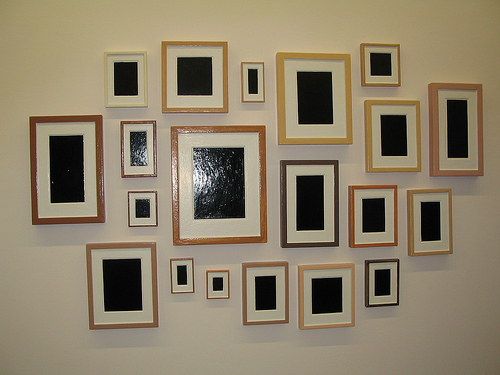}\imcreds{marcdroberts} \\
\ig{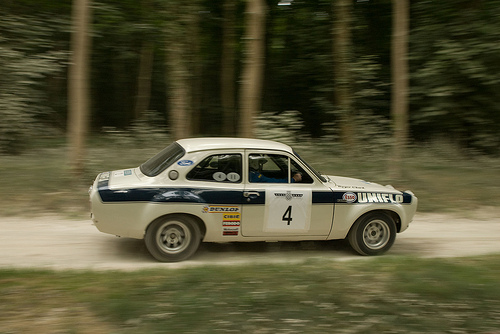}\imcreds{Greenwich Photography} & \ig{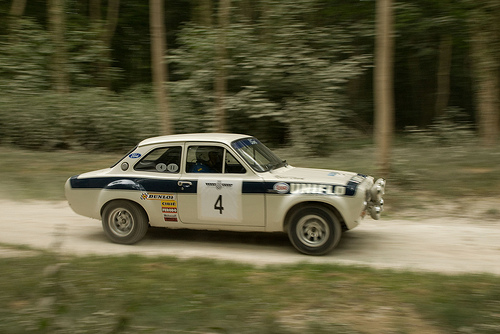}\imcreds{Greenwich Photography} \\
\multicolumn{2}{c}{pairs with 75\% of points that match} \\
\ig{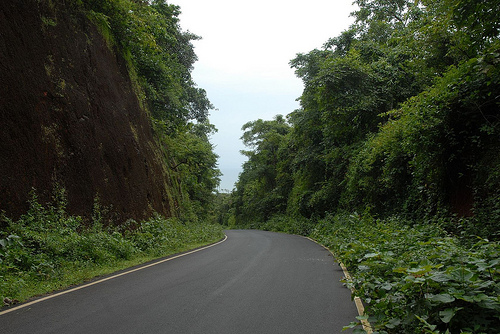}\imcreds{s\"{o}ren2013} & \ig{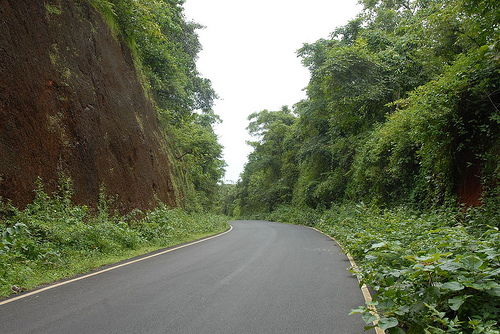}\imcreds{s\"{o}ren2013} \\
\ig{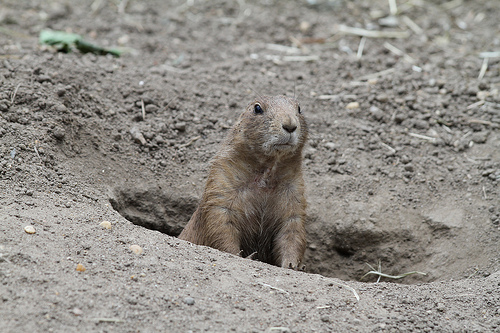}\imcreds{Jim, the Photographer} & \ig{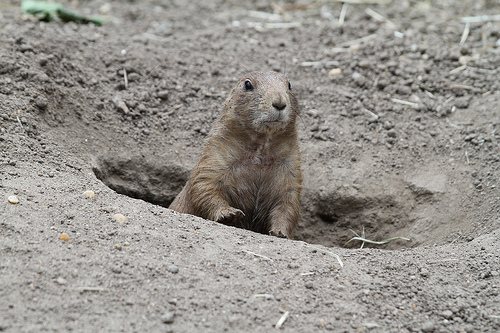}\imcreds{Jim, the Photographer} \\
\end{tabular}
    \caption{
    Examples of image pairs from the dataset. 
    }
    \label{fig:examplematches}
\end{figure}

\begin{figure}
    \centering
    \includegraphics[width=\columnwidth]{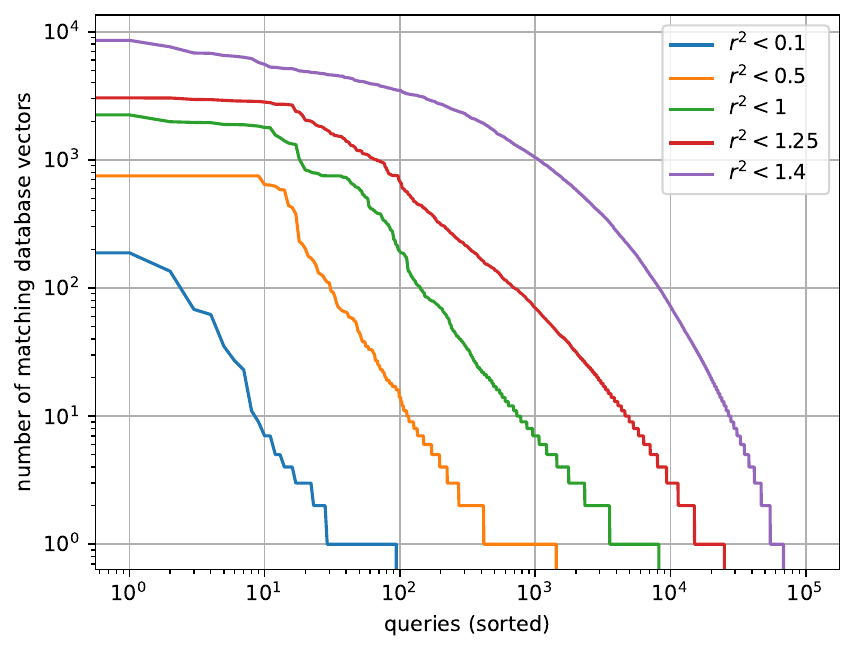}
    \caption{
    Distribution of the number of matching database vectors per query vector. 
    The queries are sorted per decreasing number of results. 
    }
    \label{fig:resultsperquery}
\end{figure}

We start from the YFCC100M dataset~\cite{Thomee2016YFCC100MTN}. 
This dataset of 100M images and videos is originally sampled from the Flickr photo sharing site. 
After removing videos, unavailable files and deduplicating identical images, we are left with 85M images. 

We shuffle that set of images randomly and split it into $\nq$=100k query images, $\nb$=10M database images and 75M training images, to avoid biases due to the order of images~\cite{Baranchuk2023dedrift}. 
We call this dataset \OURD to emphasize we do not hand-pick images to create a matching dataset.

\subsection{Image embeddings}

To perform vector search on these images, we extract 512-dimensional SSCD descriptors~\cite{pizzi2022sscd} from them. 
SSCD is as ResNet50 architecture~\cite{he16resnet} that is trained in a self-supervised way to perform image copy detection, when images undergo alterations.

The SSCD training loss includes an entropy loss term~\cite{beirlant97entropy} that is intended to ``push'' far apart images that are not matching but that otherwise would be nearby in descriptor space. 
A side-effect of this loss term~\cite{sablayrolles2019spreading} is to normalize the distance between matching and non-matching to be comparable across different queries. 

The SSCD descriptors are designed to be compared in cosine similarity. 
In this paper, we compare the vectors with squared L2 distances, which is equivalent since SSCD descriptors are L2-normalized.
\subsection{Geometrical matching}

We implement a two-stage retrieval system where, after the first-level vector search, we apply a geometrical verification.
The second level filter needs to access the image content so it is inherently more expensive. 
In this work we use a classical keypoint based image for geometrical verification. 
We extract KAZE keypoints~\cite{alcantarilla2012kaze} from the two images to compare, match them using L2 distances and run RANSAC~\cite{fischler1981random} to estimate a 4-parameter similarity transformation. 
We use OpenCV's implementation~\cite{opencv_library}.
If this estimation succeeds, the hard decision about whether the filtering passes is based on 
$\nmatch$, the number of pairs of keypoints that match and
$\ninlier$, the number of these matches that follow the geometrical model. 
We consider two settings: 
\begin{itemize}
    \item strict setting: $\nmatch \ge 10$ and $\ninlier / \nmatch \ge 0.75$. 
    \item relaxed setting: $\nmatch \ge 10$ and $\ninlier / \nmatch \ge 0.25$. 
\end{itemize}
The strict setting produces fewer, higher-confidence image matches, the relaxed setting accepts more matches. 
Our reference machine can perform 85 geometrical matching operations per second. 

Figure~\ref{fig:examplematches} shows a few examples of image pairs. 
At 9 matching points, the images are not considered matching. 
Beyond 10 matching points, with 25\% matches (the relaxed setting), the images are already clearly similar. 
With 75\% (the strict setting) the images seem to be nearby frames from the same video.

\subsection{Discussion}

The YFCC100M dataset is uncurated.
As such, it contains long series of very similar images such as video surveillance frames or template images from games, with small variations.

To visualize this, we plot in Figure~\ref{fig:resultsperquery} the number of matching image pairs in \OURD using the SSCD descriptors. 
The number of results for each query vector varies wildly. 
For example, for a search radius of 1, about 10 queries have more than 2000 matches, 1000 queries that have more than 80 matches, and 9200 (92\%) have no match at all. 

This setting where there is a power distribution of the clusters of similar images is common in collections handled by photo sharing sites or social media.
This is in contrast with datasets designed to benchmark image matching tasks. 
For example, the UKBench dataset~\cite{nister2006scalable} contains groups of exactly 4 matching images and the DISC21 dataset for image copy detection contains 0 or 1 matching database image per query~\cite{douze2021isc}.

\paragraph{No ground truth.}

This dataset does not contain ground-truth matching image pairs. 
Therefore, we use the geometrical filter as an oracle to decide whether there is a match or not.
Defining which images match is a complex task that depends on the granularity level~\cite{douze2021isc}. 

In this work, we do not question the accuracy of the oracle but instead use two different settings (strict and relaxed) to show that our observations are valid independently of the actual post-filter. 
Therefore, in the following we call pairs of images that pass the geometrical verification ``positives''.  

Performing an exhaustive geometrical matching on the $\nq\times\nb=10^{12}$ image pairs would take 373 machine-years, so it is not feasible. 
Besides, in practice the geometrical matching oracle has failure cases letting through false positives that are independent of the false positives of the pre-filtering step. 
Therefore, the matching results form the two-stage search are actually \emph{better} than with an exhaustive geometrical verification.

Therefore, this oracle enables us to measure the precision, but not the recall, as it is practically impossible to figure out how many positives pairs there are in total in the dataset.
In the following section, we design a metric that does not need the recall and is efficient to evaluate.

%% file: 3-metric.tex
In this section, we start by describing the post-filtering setup. 
We show that counting-based methods are bound to give noisy metrics. 
We then we introduce the RSM as an expectation of a number of positives for a given filtering budget.

\subsection{Vector search as a pre-filter}
\label{sec:prefilter}

We consider that the vector search is part of a multi-stage retrieval system where vector search is the fast, but inaccurate first retrieval step. 
The results from this first stage are post-filtered by some more costly post-verification process. 
We are given a fixed budget $B$ for this operation.
The budget models real-world constraints on compute. 
This post-filtering could be a manual verification, or another costly operation. Only the (query, database) pairs that pass the filter are useful for the application, \ie they are positives.

For the \OURD dataset, we use the geometrical verification as the post filter. 
In the following, we set the budget to $B=10^5$ for the strict setting and $B=10^6$ for the relaxed setting, that lets through more matches by design. 

\paragraph{Bulk filtering.}

When queries are considered one by one, the most relevant elements to post-verify are the top-$B$ retrieved items. 

When a set of $\nq$ queries is provided at once,  there are several ways of ``spending'' the verification budget. 
The first way is by re-ranking the $B/\nq$ first search results for each query. 
This shortlist is built with a k-nearest-neighbor search on the vectors ($k=B/\nq$). 

The second approach is to set a global distance threshold $r$ calibrated to return $B$ results. 
In that case, the shortlist is built with range search. 

\begin{figure}
    \centering
    \includegraphics[width=\columnwidth]{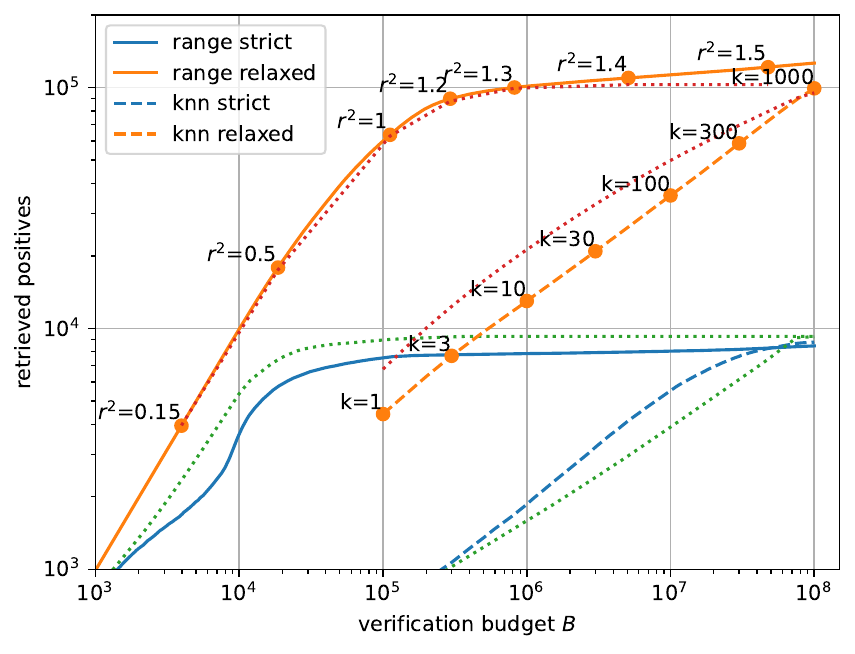}
    \caption{
Number of positives found after the filtering stage (with the strict and relaxed settings) for two types of vector search: range search and k nearest neighbor (knn) search.     
For some points, we indicate the search radius ($r^2$, for range search) and the number of results per query ($k$, for knn search). 
The dotted lines are the estimated counts based on the RSM.
}
    \label{fig:verificationbudget}
\end{figure}

Figure~\ref{fig:verificationbudget} shows that in the relaxed setting, and range search, retrieving 100,000 positive pairs is relatively efficient (10\% of the shortlist are positives).
Beyond these, reaching more positive pairs becomes exponentially more expensive. 

The k-NN search, that verifies a fixed number of results per query is comparatively less efficient, except possibly for very high budgets: for $B=10^8$ the filter pass rate is around 0.1\% for the relaxed setting or 0.01\% for the strict one. 
Therefore, for practical applications,  range search is the most appropriate approach and it is useful in the small radius regime. 
Thus, approximate search methods need to focus on that part of the distribution.

\subsection{Distance to neighbors distribution}

To gauge what happens with short radius search, we study the distribution of distances to neighbors within a small radius. 
The distance distribution can be derived in closed form for some simple continuous vector distributions. 
In the following, both the query vector $q$ and the database vectors $x$ are drawn independently from the same distribution. 

\paragraph{Uniform vectors on the unit sphere.}

Without loss of generality, we consider the distance between unit vector $q=(1, 0, ..., 0) \in \mathbb{R}^d$ and the other vectors~\cite{iscen2017memory}. 

The set $S(r)$ of points at distance $r$ from $q$ is the intersection of the $0_d$-centered hypersphere of radius 1 and the $q$-centered hyphersphere of radius $r$. 
This is a $d-1$ dimensional hypersphere of radius $h$ that lays on the hyperplane of constant first coordinate $x_1$ that verifies: 
\begin{equation}
r^2 = (1-x_1)^2 + h^2 \textrm{  and }
1 = h^2 + x_1^2    
\end{equation}
Which yields $x_1 = 1-r^2/2$ and $h=r\sqrt{1-r^2/4}$. 
The surface of $S(r)$ is 
\begin{equation}
\mathcal{S}(r) = 
     C_{d-2} h^{d-2} = 
     C_{d-2} r^{d-2} \left(1-\frac{r^2}{4}\right)^\frac{d-2}{2}
\end{equation}
with $C_{d-2}$ a factor that only depends on the dimension.
The angle between $q$ and $S(r)$ is constant: $\beta = \cos^{-1}(1-r^2/2)$. 
When $r$ is increased by an infinitesimal step $\partial r$, $S(r)$ sweeps a hypercone trunk of length $\partial \beta$, where 
\begin{equation}
    \frac{\partial \beta}{\partial r} = 
    \frac{2r}{\sqrt{4-(2-r^2)^2}}
\end{equation}
The surface of this hypercone trunk is the product $\partial \beta \times \mathcal{S}(r)$. 
Thus, the density function of the distances is 
\begin{equation}
    \phi(r) \sim 
    \frac{r^{d-1}}{\sqrt{4-(2-r^2)^2}}
    \left(1-\frac{r^2}{4}\right)^\frac{d-2}{2} 
    \label{eq:pdfuniform}
\end{equation}

\paragraph{Gaussian vectors.}

We consider the simplest case where the query and database vectors come from a unit normal distribution: $q, x \sim \mathcal{N}(0_d, \frac{1}{2} \mathrm{Id}_d)$. 

The difference $y=q-x$ follows the same distribution as the sum $q + x \sim \mathcal{N}(0_d, \mathrm{Id}_d)$ because $q$ and $x$ are drawn independently. 
The norm of the difference $y$ can be decomposed over dimensions:  
\begin{equation}
   \|q - x\|^2 = \|y\|^2 = \sum_{i=1}^d y_i^2
\end{equation}
Since the covariance is diagonal, the $y_i$'s are i.i.d.,  
their distribution is Gaussian with variance $\sigma=1$. 
The sum of $d$ independent squared Gaussian variables follows a Chi-squared distribution with $d$ degrees of freedom, which gives
$
    \|y\|^2 \sim \chi^2(d)
$.
Noting $x \mapsto C'_d x^{d/2-1}e^{-x/2}$ the pdf of the Chi-squared distribution, with $C'_d$ a term that depends only on the dimensionality, then 
the density function is: 
\begin{equation}
    \phi(r) \sim r^{d-1} e^{-r^2/2}
    \label{eq:pdfgaussian}
\end{equation}

\paragraph{Discussion.}

Interestingly, for the densities in Equations~(\ref{eq:pdfuniform}) and (\ref{eq:pdfgaussian}), the first non-0 terms of the Taylor expansion is $r^{d-1}$. 

\begin{figure}
    \centering
\includegraphics[width=\columnwidth]{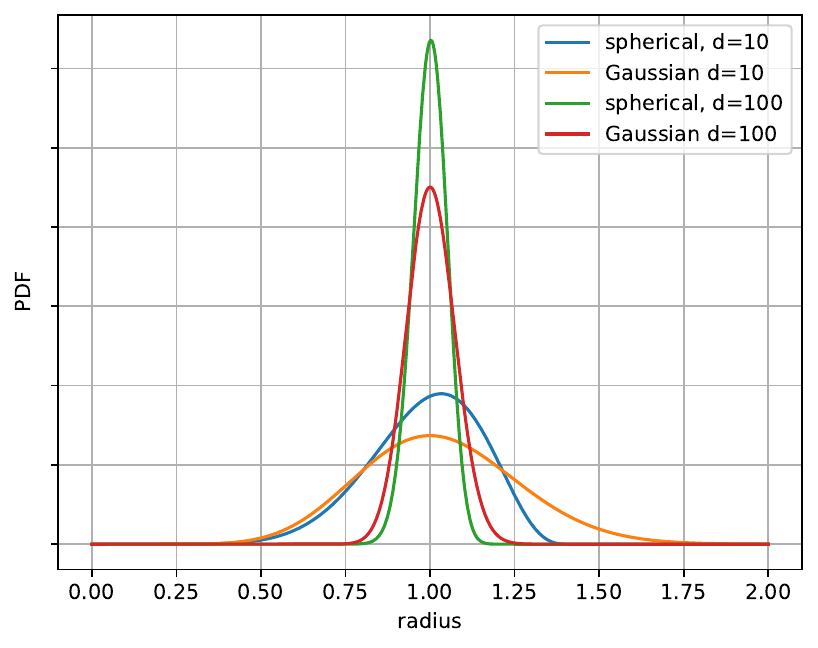}
    \caption{
Distribution of distances between points for a uniform spherical distribution and a Gaussian distribution in dimensions $d=10$ and 100. 
The vector distributions are scaled so that the mode of each distance distribution is at 1. 
    }
    \label{fig:densities}
\end{figure}

Since the dimension $d$ is 10s to 100s, this means that the increase of the number of neighbors is very sudden, see Figure~\ref{fig:densities}.
This is also what we observe in practice.  

The consequence is that when searching with a fixed small radius, most results are close to the range boundary.
Metrics like precision and recall are based on counting vectors that fall inside/outside the boundary. 
Therefore, for a given query, these metrics depend disproportionately on the matches near the range boundary, while closer points are 
(1) more stable when affected by a small amount of noise and 
(2) more important for the end application, because they are less likely to be filtered out. 
In the following, we propose a model of the utility of a search result. 

\subsection{Probability of being filtered out}

At the basis of the RSM, there is a model of the expected filter passing rate given the measured vector distance. 

We model the probability for a pair $(q, x)$ of a query and a database point to be positive for the application for a given distance: 
\begin{equation}
    f(r) = P[(q, x) \textrm{ positive } | \|q - x\| = r]
\label{eq:probatp}
\end{equation}
The function $f$ is monotonically decreasing and dropping to 0 \ie it is less likely for farther apart vectors to be positive matches. 
We are going to build upon $f$ as the value function of a search result to define the range search metric. 
This assumes the feature extraction is trained with some form of distance calibration in its loss. 
Given the filtering rate model, we can estimate the number of verified items from $f$ and measured distances.

\subsection{The Range Search Metric (RSM)}

\paragraph{Single query.}

Given a query vector $q$, and a verification shortlist $S=\{c_1, c_2, ..., c_B\}$, then the expected number of positives is: 
\begin{equation}
    E[\textrm{\# positives in } S] = \sum_{i=1}^n f(\|q - x_{c_i}\|^2)
\label{eq:expectedpos}
\end{equation}

Since $f$ is a decreasing function, for a given verification budget $B$, returning the top-$B$ nearest database vectors as $S$ maximizes the expected number of positives. 
This is a k-nearest neighbor search operation. 

\paragraph{Multiple queries.}

Given a set of $\nq$ queries, $\{q_1,..., q_{\nq}\}$, $B$ is the total verification budget and $B_i$ is the number of results returned for query $i$: $B=\sum_{i=1}^{\nq} B_i$.
The total expected positives is our \textbf{Range Search Metric (RSM)}: 
\begin{equation}
\mathrm{RSM} = %
\sum_{i=1}^{\nq} \sum_{j=1}^{B_i} f(\|q_i - x_{c_{ij}}\|^2)
\label{eq:expectmultiple}
\end{equation}
where $c_{ij}$ is the $j$'th element in query $i$'s list of results. 

To maximize the RSM, we should keep the top-$B$ smallest distances between pairs of vectors. 
This is equivalent to setting a global threshold on distances and keeping all pairs of points below that threshold: this is the range search operation. 

This explains the bulk filtering result of Figure~\ref{fig:verificationbudget}, where k-NN search is less efficient than range search to find neighbors.

\paragraph{Approximate search.}

For approximate search, Equation~(\ref{eq:expectmultiple}) estimates the number of positives in the same way, except that the $c_{ij}$ are the approximate search results. 

The advantage of using the RSM metric rather than directly performing the geometric verification is that it is faster to evaluate. 
In Section~\ref{sec:method}, we measure the impact of approximate search on the RSM metric, it would be infeasible to repeat the experiment of Figure~\ref{fig:verificationbudget} for multiple indexes.

\begin{figure}
    \centering
    \includegraphics[width=\columnwidth]{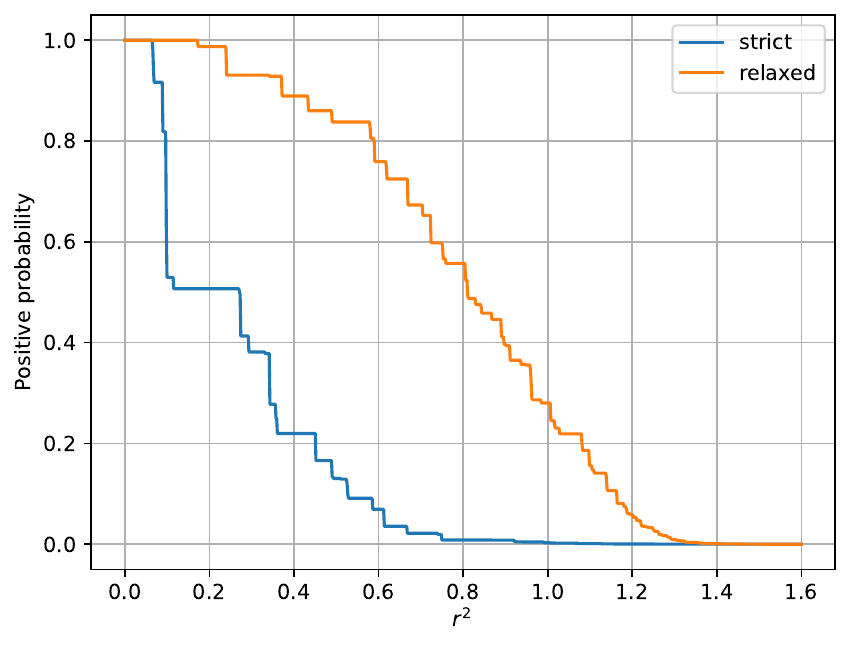}    
    \caption{
    Estimated positive probability for the \OURD dataset. 
    The positive probabilities are obtained by isotonic regression. 
        }
    \label{fig:isotonic}
\end{figure}

\subsection{Model of the positive function}

Given a representative sample of positives and negatives pairs of vectors, it is possible to model the probability of Equation~(\ref{eq:probatp}) using isotonic regression. 

Positive query results are represented as $(\|q-x\|^2, 1)$ and negative results as $(\|q-x\|^2, 0)$, yielding $(X_i, Y_i)_{i=1..N}$ input/output pairs, ordered by increasing distance $X_i$. 
The isotonic regression consists in solving 
\begin{equation}
    \underset{\widehat{Y}_1 \ge ... \ge \widehat{Y}_N}{
    \mathrm{Minimize}}
    \sum_{i=1}^{N} (\widehat{Y}_i - Y_i)^2
    \label{eq:isotonic}
\end{equation}
Thus, the $\widehat{Y}_i$'s are probabilities associated to each existing distance between pairs of vectors. 
The function $f$ is then defined at coordinates $X_i$: $f(X_i) = \widehat{Y}_i$. 
The intermediate values can be defined \eg by linear interpolation.

Equation~(\ref{eq:isotonic}) can be re-stated as a quadratic programming problem that is solved in linear time with the iterative algorithm of pool adjacent violators~\cite{best1990active}. 
We use scikit-learn's isotonic regression implementation~\cite{scikit-learn}.

To justify the formalization of Equation~(\ref{eq:isotonic}), we can examine what happens for a set of equal distances $X_n=X_{n+1}=..=X_m$, with positive and negative results $(Y_i)_{i=n..m}$. 
The empirical probability of positives is $(\sum_{i=n}^{m} Y_i) / (m - n + 1)$. 
This is also the solution to $\mathrm{argmin}_{\widehat{Y}} \sum_{i=n}^m (\widehat{Y} - Y_i)^2$, which coincides with the terms $n,...,m$ of Equation~(\ref{eq:isotonic}).
Therefore, for this particular case, the isotonic estimation matches the empirical probability.

\paragraph{Application to \OURD.}

Figure~\ref{fig:isotonic} shows the regression results for the \OURD dataset, trained on vector pairs extracted from the training set (100k queries, 1M database vectors).
The model is piecewise linear, it drops to almost 0 beyond a squared radius of 1.4, even earlier for the strict setting. 
Figure~\ref{fig:verificationbudget} shows the RSM estimation of the number of pairs matches. 
It matches the measured pairs of images that pass the filter quite accurately. 
In the following, we use the RSM metric to benchmark approximate search techniques.

%% file: 4-method.tex
In this section we evaluate how typical vector search indexes perform in light of the RSM metric. 
We focus on the most scalable indexes, that rely on a of the vectors and on vector compression.

\subsection{Coarse quantization}

\begin{figure}
    \centering
    \includegraphics[width=\columnwidth]{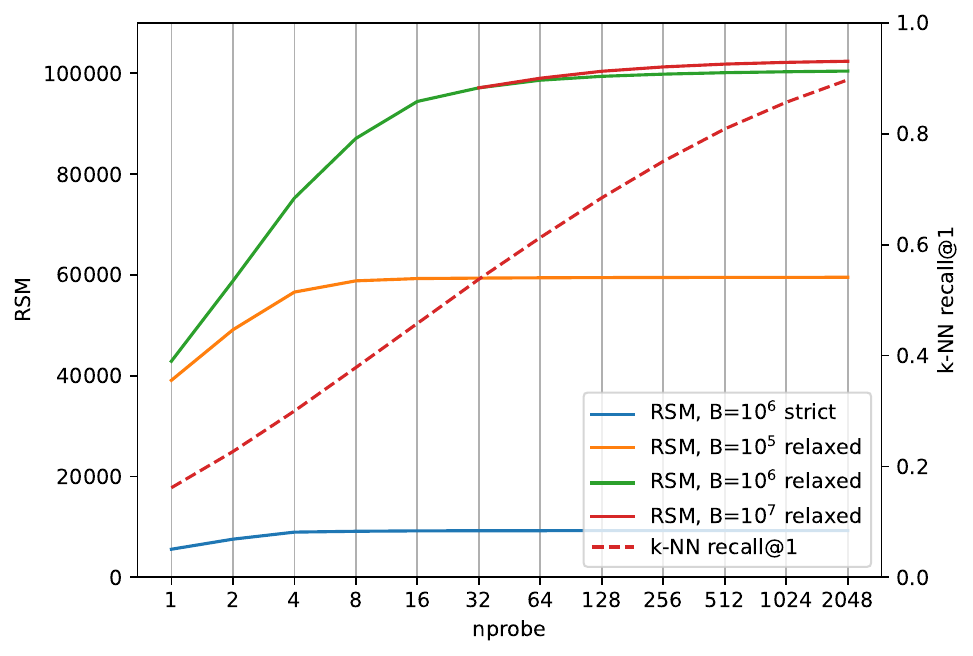}
    \caption{
Indexing \OURD with an index of $2^{16}$ clusters, without compression (\texttt{IVF65536,Flat} index in FAISS terms). 
We compare metrics for k-NN search (recall) and for range search (RSM).     
    }
    \label{fig:RSMvsknn}
\end{figure}

\paragraph{Comparison with k-NN search.}

The first experiment evaluates the performance of the simplest IVF index type, the IVFFlat~\cite{douze2024faiss}.
The database vectors are clustered with k-means. 
At search time, the \verb|nprobe| nearest clusters to the query vector are visited and the vectors they contain are compared with the query. 

Figure~\ref{fig:RSMvsknn} shows the results. 
The \verb|nprobe| parameter adjusts the speed vs. accuracy tradeoff: search time is roughly proportional to it, with a fixed overhead.
The k-NN search is evaluated with the nearest neighbor recall, which is typical. 
The results show that k-NN search benefits from increasing the \verb|nprobe| parameter. 
This is much less the case for range search: the RSM saturates after \verb|nprobe|=8 for the strict setting or verification budgets $B\le 10^5$. 
For higher budgets in the relaxed setting, the RSM does not improve beyond \verb|nprobe|=256, which visits less than 0.4\% of the clusters.
This means that only the closest neighbors do matter for the RSM metric.

\paragraph{Fast coarse quantizers.}

\begin{figure}
    \centering
    \includegraphics[width=\columnwidth]{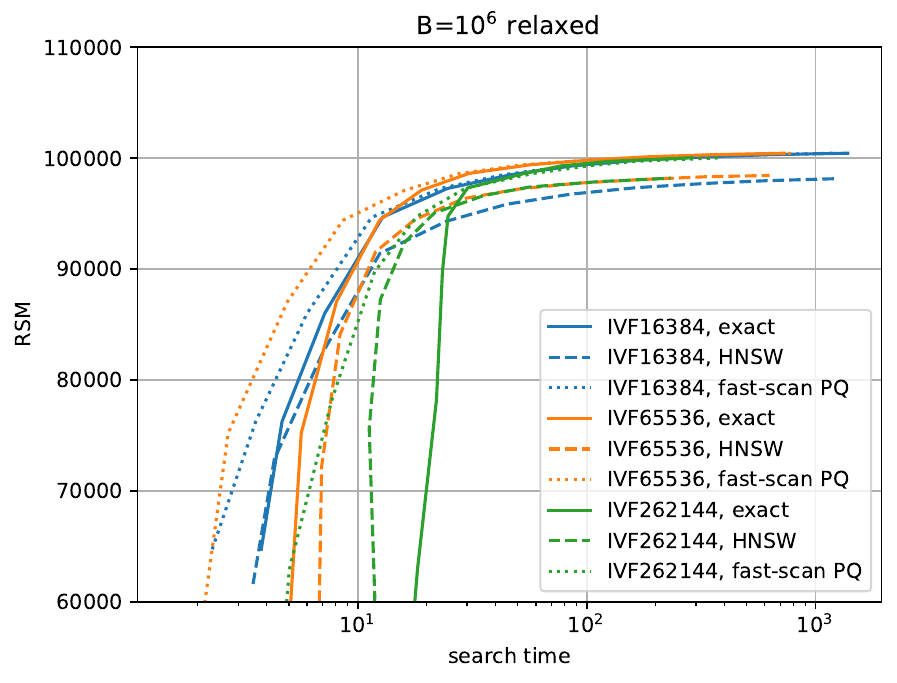}
    \caption{
Comparison of approximate clustering variants for the RSM metric.    
IVF$n$ means there are $n$ clusters, we compare three variants to perform the cluster assignment: exact search, search with HNSW and an optimized (fast-scan) version of product quantization (PQ). 
    }
    \label{fig:RSMcoarsequantizer}
\end{figure}

The assignment to IVF clusters uses a vector search that finds the nearest centroids to the query vector. 
This vector search can be exhaustive (like in the previous paragraph) but for large indexes it is more efficient to use an approximate search. 
Figure~\ref{fig:RSMcoarsequantizer} compares various settings for the coarse quantizer. 
It shows that the best setting is with 65536 clusters. 
To do approximate search, fast-scan PQ is preferable~\cite{andre2019quicker}, it performs the search with a first-level filtering based on product quantization, then follows with an exact re-ranking. 
The graph-based HNSW's accuracy saturates in the higher RSM regime because it ``misses'' some centroids. 

\subsection{Vector codec}

\begin{table*}
\centering
    \begin{tabular}{|ll|rr|rr||rr|}
    \hline
Code  & 
Encoding $\downarrow$ & 
\multicolumn{2}{c|}{strict, $B=10^5$} &
\multicolumn{2}{c||}{relaxed, $B=10^6$} & 
\multicolumn{2}{c|}{k-NN recall\@1}\\
size $\downarrow$& by residual & no & yes & no & yes & no & yes \\
\hline
8 & ITQ64    & 6.78 &    0.16 &   29.30 &    0.91 &   0.023 &   0.004\\
8 & PQ16x4   & 0.06 &    8.23 &    2.83 &   71.47 &   0.040 &   0.060\\
8 & PQ8x8    & 6.86 &    {\bf 8.26} &   44.03 &   {\bf 72.18} &   0.045 &   {\bf 0.067}\\
\hline
16 & ITQ128   & 8.36 &    0.50 &   56.95 &    2.58 &   0.053 &   0.015\\
16 & PQ32x4   & 8.14 &    8.54 &   48.51 &   74.84 &   0.087 &   0.104\\
16 & PQ16x8   & 8.37 &    {\bf 8.64} &   61.77 &   {\bf 75.41} &   0.096 &   {\bf 0.113}\\
\hline
32 & ITQ256   & 8.83 &    0.70 &   75.97 &    3.62 &   0.098 &   0.025\\
32 & PQ64x4   & {\bf 8.91} &    8.80 &   83.72 &   77.87 &   0.167 &   0.186\\
32 & PQ32x8   & 8.90 &    8.80 &   {\bf 85.38} &   78.17 &   0.184 &   {\bf 0.199}\\
\hline
64 & ITQ512   & 8.92 &    1.07 &   90.37 &    8.66 &   0.186 &   0.081\\
64 & PQ128x4  & 8.95 &    8.92 &   94.95 &   89.59 &   0.292 &   0.286 \\
64 & PQ64x8   & {\bf 8.95} &    8.94 &   {\bf 95.31} &   92.01 &   {\bf 0.303} &   0.296\\
\hline
2048 & Flat & 8.96 & 8.96 &  99.40 & 99.40 &  0.685 & 0.685  \\
\hline
    \end{tabular}
\caption{
    RMS ($\times 1000$) measures for an IVF index with 65536 centroids and different encodings. 
    ITQ$n$ means an ITQ code of $n$ bits. 
    PQ$n$x$m$ is a product quantizer encoding with $n$ groups of $m$ bits. 
    }
\label{tab:encodings}
\end{table*}

At search time, the partitioning performed by the coarse quantizer selects a subset of vectors that need to be compared with the query vector. 
However, since storage is in limited supply, the vectors are stored in a compressed format. 
The two compression formats of interest are ITQ and PQ. 
ITQ produces binary representations~\cite{jegou2008hamming,gong2013iterative} that are compared with Hamming distances (symmetric distance computation). 
PQ splits input vectors into sub-vectors, and separately encodes each sub-vector with a vector quantizer~\cite{jegou11pq}. 
At search time, the PQ distances are asymmetric, \ie the exact query representation is compared with an approximate database vector. 

For PQ we report two variants: one where each sub-vector is encoded in 8 bits and one for 4 bits. 
The second one has a more efficient SIMD search implementation~\cite{douze2024faiss}.

For the vector codec experiments, we fix the IVF setting at 65536 centroids  with exact assignment. 
We compare the RMS measure for the strict setting with $B=10^5$ and the relaxed setting with $B=10^6$, and report the k-NN results for reference.

Table~\ref{tab:encodings} shows the results, with the exact representation for reference (``Flat''). 
For all metrics, the residual encoding is beneficial for short codes (less than 16 bytes), as observed before for k-NN search~\cite{douze2024faiss}.

For k-NN search, the recall increases steadily with the size of the codes, from 0.067 at 8 bytes to 0.296 at 64 bytes. 
However, this accuracy is sill way below that of the exact representation (0.685). 
For a given size, it is more accurate to allocate more bits to fewer dimension groups~\cite{amara2022nearest}, \ie PQ8x8 is more accurate than PQ16x4, which is more accurate than ITQ64. 
This is clearly visible in the k-NN recall results. 

In contrast, the RSM metric tends to saturate with the code sizes and for variants of the same code size: 
for example, the gap between ITQ256 and PQ32x8 is a factor 8 for k-NN search while its impact is 12\% relative for RMS in the relaxed setting. 
This is even more true for the strict setting. 
Besides, in this setting, the search accuracy mostly saturates beyond 16 bytes. 

This is because in that setting the matching relies on ``obvious'' matching images that are already identified by small representations. 

\subsection{Discussion}

The major observations of interest for range search are that it is important to have an accurate coarse quantizer (fast-scan PQ). 

Compared to k-NN search, it is less important to have the most accurate vector quantizer,
\eg using a binary representation, while slightly suboptimal, makes distance comparisons 6$\times$ faster than an 8-bit PQ. 
Besides, when the post-filtering is strict, it is not necessary to use very large encodings for the vectors.